 \documentclass[a4paper,12pt]{amsart}
 \usepackage{amsmath,amsthm,amssymb}
 \usepackage{graphicx,ascmac}
 \usepackage{txfonts}
\usepackage{hyperref}
\usepackage[margin=2cm]{geometry}

\theoremstyle{definition}
\newtheorem{thm}{Theorem}[section]

\newtheorem{rem}[thm]{\rm Remark}
\newtheorem{ex}[thm]{\rm Example}
\numberwithin{equation}{section}

\newcommand{\R}{\mathbb{R}}

\newcommand{\Z}{\mathbb{Z}}

\usepackage{listings}
\makeatletter
\@namedef{subjclassname@2020}{%
  \textup{2020} Mathematics Subject Classification}
\makeatother

\title{Cubical Ripser: Software for computing persistent homology of image and volume data}

\author[S.~Kaji]{Shizuo KAJI}
\address[S.~Kaji]{Institute of Mathematics for Industry, Kyushu University, 
744 Motooka, Nishi-ku, Fukuoka, 819-0395, Japan}
\email{skaji@imi.kyushu-u.ac.jp}

\author[T.~Sudo]{Takeki Sudo}
\address[T.~Sudo]{}
\email{takeki.sudo@gmail.com}

\author[K.~Ahara]{Kazushi Ahara}
\address[K.~Ahara]{School of Interdisciplinary Mathematical Sciences, Meiji University, 4-21-1, Nakano, Nakano-ku, Tokyo, 164-8525, Japan}
\email{ahara@meiji.ac.jp}

\date{\today}

\begin{document}
\maketitle

\begin{abstract}
We introduce Cubical Ripser for computing persistent homology of image and volume data
 (more precisely, weighted cubical complexes).
To our best knowledge, Cubical Ripser is currently the fastest and the most memory-efficient program for computing persistent homology of weighted cubical complexes.
We demonstrate our software with an example of image analysis in which persistent homology and convolutional neural networks are
successfully combined.
Our open-source implementation is available at \cite{github}.
\end{abstract}

\section{Introduction}
Recent years, Topological Data Analysis (TDA, for short) has gained much attention as a new way of looking at data
(\cite{carlsson,survey} provide a survey).
One of the main computational tools of TDA is persistent homology, which extracts topological features
 from weighted cell complexes.
Point clouds, time series, and images can be turned into weighted cell complexes.
Extracted features of data by persistent homology are used as input for statistical techniques 
to provide new insights on the data which are overlooked with conventional feature extractors (see, e.g., \cite{Hiraoka,McGuirl}).

Most of the conventional feature extractors for images focus on \emph{local} information of images.
For example, convolution looks at the relation among neighbouring pixels.
On the other hand, homology provides a way to encode \emph{global} information of images 
into numeric values. 
Global features are what humans usually perceive from data and are generally robust.
Persistent homology provides efficient machinery to extract global features of data.
Those features obtained through persistent homology can be used in conjunction with conventional methods.
For example, a hybrid approach to image analysis combining persistent homology and deep learning was proposed in \cite{DL}.

There are various pieces of software to compute persistent homology; DIPHA, GUDHI, and Ripser, to name a few popular ones.
Ripser \cite{Ripser} is known to be the most efficient in terms of memory and time 
for computing persistent homology of the Vietoris-Rips complexes.
The Vietoris-Rips complex is one of the major forms of simplicial complex arising from point clouds.
Ripser cannot be directly applied to other types of data, including images,
which are usually modelled by weighted cubical complexes.

In this note, we introduce Cubical Ripser for computing persistent homology of weighted cubical complexes
 (with the coefficients in $\Z/2\Z$),
which was initially developed during the master's program of the second author under the supervision of the third author \cite{Sudo}.
A modification including easy-to-use python bindings for the 2-dimensional version was created by Carrara \cite{cube2d}.
The current version of Cubical Ripser computes persistent homology of 1D, 2D, and 3D arrays of floating-point values, which include 
scalar time-series (1D), greyscale pictures (2D), and volume data (3D).
The core algorithm used in Cubical Ripser is an adaptation of that of Ripser 
to cubical complexes and inherits efficiency from its parent.
The primary goal of this note is to provide practitioners with quick access to the tool.
In particular, we introduce a novel method to incorporate 
both local and global information of persistent homology 
to deep learning (Example \ref{ex:NN}).

Here, we give an intuitive account of the persistent homology of 2D/3D images.
Given a greyscale image,
we can associate a binary image consisting of those pixels with values less than or equal to $a$, where $a\in \R$ is a real number. 
This is nothing but thresholding. We call the resulting binary image \emph{the sublevel set} at $a$.
From this binary image, we can extract its topology by the usual notion of homology; 
by counting the number of connected components, the number of loops, and the number of voids.
Persistent homology combines the homology of the sublevel sets by sweeping the threshold through the whole real numbers.
The information of persistent homology is expressed as \emph{barcodes}, a set of intervals in the real numbers.
More specifically, Cubical Ripser takes images as input and outputs a set of pairs of the real numbers.
Each pair is denoted by $[a,b) \ (a,b\in \R\cup \{\infty\})$
and is either of dimension 0,1, or 2. 
A $0$-dimensional pair $[a,b)$ represents a connected component which emerges at threshold $a$ and 
disappears at $b$ (see Figure \ref{fig:H0}).
A $1$-dimensional (resp. $2$-dimensional) pair represents a loop (resp. void) in a similar manner.
A $d$-dimensional pair $[a,b)$ is also referred to as a $d$-\emph{cycle with lifetime} $b-a$.
Features in the form of barcodes may not be very convenient for further processes.
There are techniques to give a concise vector representation for barcodes such as
the persistence images \cite{persistence_images} so that persistent homology fits in typical data-analysis pipelines.


\section{Persistent homology of images}
Let us start with a formal definition of the object we deal with in this note.
An \emph{image} is a real-valued function over a regular grid.
A \emph{3-dimensional regular grid} of size $N$ is the set $\Omega^3_N=\{ (i,j,k) \in \Z^3 \mid 0\le i,j,k < N \}$,
and a \emph{3-dimensional image} (or \emph{volume}) is a function $\phi: X \to \R$, where $X$ is a subset of $\Omega^3_N$.
A 2-dimensional image is a special case when $X$ is contained in the subspace $k=0$.
Similarly, 
a 1-dimensional image is a special case when $X$ is contained in the subspace $k=j=0$.

Typical examples include greyscale pictures, CT and MR images, and AFM.
Note that in our definition, images can be defined on a non-rectangular region such as a disk.
In this case, we can extend $\phi$ to the whole $\Omega^3_N$ by assigning $\infty$ 
(in practice, a predefined huge floating-point number) outside the domain.

\begin{figure}[ht]
\center
\includegraphics[width=13cm]{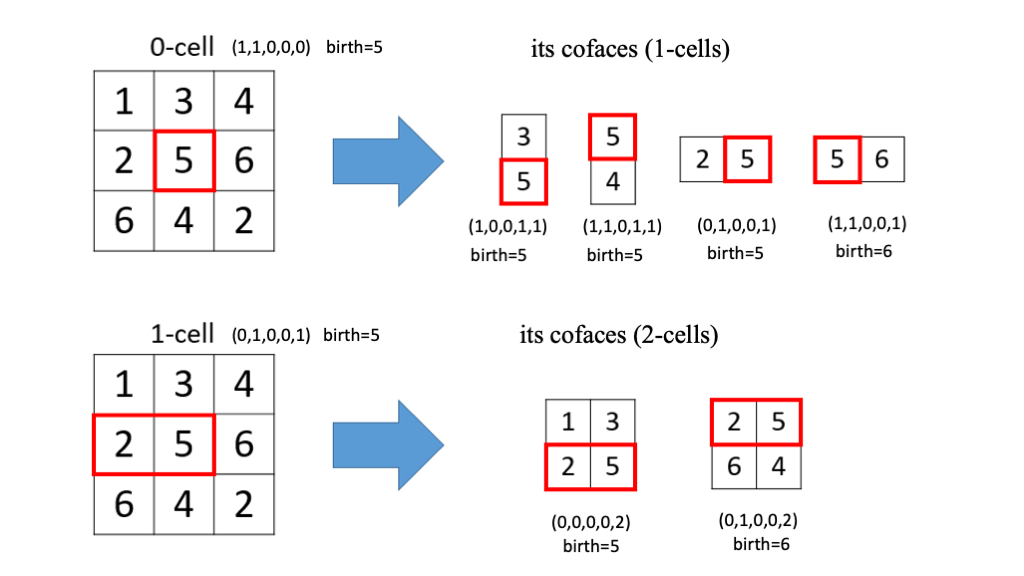}
\caption{0- and 1-cells and their cofaces in a 2-dimensional image.
The numbers in the boxes indicates the function value.
Our convention is that the coordinates of the upper-left voxel is $(0,0,0)$
and that of the cell to the right is $(1,0,0)$.}\label{fig:coface}
\end{figure}

Our computation strategy for persistent homology is the same as that of Ripser;
computing cohomology by matrix reduction (\cite{cohomology}) with 
 implicit representation of the coboundary matrix.
We refer the reader to \cite{cohomology} for the details of the algorithm.
We only give a brief account of what is peculiar to images.
For an image $\phi: X\to \R$ is the associated \emph{weighted cubical complex}
$K(\phi)$ consisting of \emph{cubical cells}.
Each cubical cell is identified with
a tuple $(x,y,z,m,d)$, where 
$(x,y,z) \in X\subset \Omega^3_N$ is the \emph{location},
$m\in \{0,1,2\}$ is the \emph{type}, and $d\in \{0,1,2,3\}$ is the dimension
 of the cell.
A 0-dimensional cell is just a pixel/voxel. There is only a single type $m=0$ in dimension $0$.
The \emph{birth-time} (or the \emph{weight}) of the cell $(x,y,z,0,0)$ is given by the 
function value $\phi(x,y,z)$.
A 1-dimensional cell $(x,y,z,0,1)$ of type $m=0$ consists of two 0-cells
$(x,y,z,0,0)$ and $(x+1,y,z,0,0)$.
A 1-dimensional cell $(x,y,z,1,1)$ of type $m=1$ consists of two 0-cells
$(x,y,z,0,0)$ and $(x,y+1,z,0,0)$.
A 1-dimensional cell $(x,y,z,2,1)$ of type $m=2$ consists of two 0-cells
$(x,y,z+1,0,0)$ and $(x,y+1,z,0,0)$.
Similarly, we have three types of $2$-dimensional cell.
There is only one type of $3$-dimensional cell.
The birth-time of a cell is defined by the maximum of the birth-time of the 0-cells contained in it.
The set of \emph{cofaces} of a $d$-cell is the set of $(d+1)$-cells containing it.
These definitions are easily understood from Figure \ref{fig:coface}.

\begin{figure}[ht]
\center
\includegraphics[width=3cm]{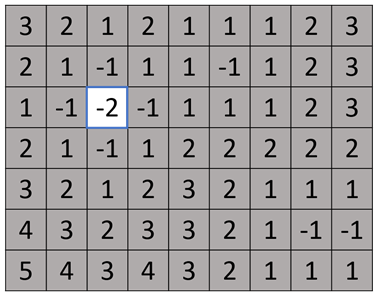}
\includegraphics[width=3cm]{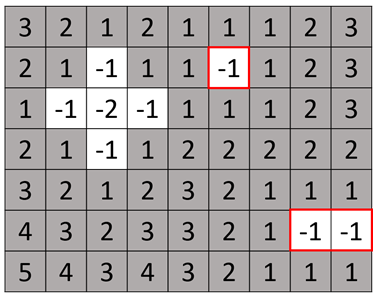}
\includegraphics[width=3cm]{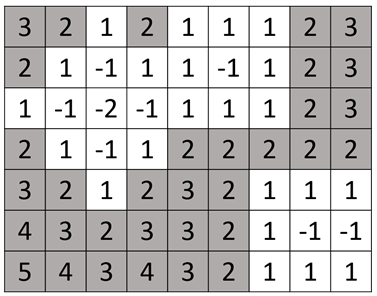}
\includegraphics[width=3cm]{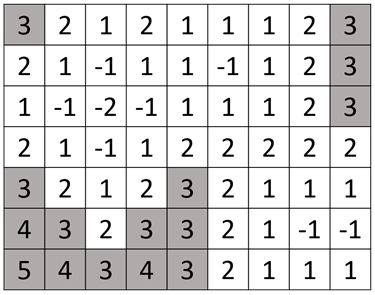}
\caption{0-th persistence of a 2D image.
From left to right, 
in the sublevel set $\phi \le -2$ a connected component is born,
in $\phi \le -1$ two other connected components are born,
in $\phi \le 1$ the upper one with birth-time $-1$ is killed and merged into the one with birth-time $-2$,
in $\phi \le 2$ the right one with birth-time $-1$ is killed.
Therefore, the 0-th barcodes are $[-2,\infty), [-1,1)$, and $[-1,2)$.
}\label{fig:H0}
\end{figure}

The 0-th persistent (co)homology is nothing but the transition of the connected components of 
the sublevel set of $\phi$. It is computed efficiently by the union-find algorithm.
Two 0-cells are connected if there is a sequence of 1-cells connecting them.
In other words, we consider 4-neighbour connectivity in 2D images and 6-neighbour connectivity in 3D images.
Traversing 1-cells in the increasing order of birth-time and 
merging connected components, we obtain the 0-th persistent (co)homology.
When two connected components are joined by a 1-cell, the one with the larger birth-time is \emph{killed}.
See Figure \ref{fig:H0} for the computation of 0-th persistence.

The 1st and 2nd persistent cohomology is computed by reducing the coboundary matrix.
We give the set of $d$-cells a total ordering by the lexicographic ordering
of $(-\text{birth-time},m,z,y,x)$; cells with larger birth-time come first, 
and among cells with the same birth-time, those with smaller type come first, etc.
We index cells in the increasing order.
The coboundary matrix $D$ is defined so that its $(i,j)$ entry is one if and only if the $j$-th cell has the $i$-th cell as a coface.
The matrix $D$ is sparse, and each column can be computed easily on-the-fly by enumerating cofaces of a cell.
What we have to do to compute persistent cohomology is to turn $D$ into upper-triangular form by column operations.
The pivots in the reduced coboundary matrix correspond to cycles:
if the $(i,j)$ entry is the pivot, it corresponds to a cycle with birth-time equal to that of the $j$-th cell
and death-time equal to that of the $i$-th cell.
Thanks to the property $D^2=0$, many columns are known to be reduced to zero without computation.
See \cite{cohomology} for details.

\begin{rem}
With this scheme, the computation proceeds from the lowest degree (degree 0) to higher degrees.
When we want to compute only the top degree persistent homology (that is, 1-cycles for 2D images and 2-cycles for 3D images),
we can directly compute it using the version of the Alexander duality discussed in \cite{duality}.
In Cubical Ripser, it is enabled by the command-line option (\verb+--top_dim+).
\end{rem}

\begin{rem}
Sorting cells could be improved by first sorting voxels (0-cells) by the birth-time.
Since the birth-time of a cell is the largest birth-time of 0-cells contained in it,
higher cells can be sorted by enumerating all the cells containing a 0-cell.
However, in our situation, we have to sort only twice; for 1-cells and for 2-cells.
So we did not use this possible optimisation in favour of the simplicity of the codes.
In practice, we observed that the optimisation gained us only less than 1\% speed up.
\end{rem}

\section{Timing}
We compared the computational efficiency of Cubical Ripser with DIPHA \cite{dipha}, 
one of the most popular and efficient software which computes persistent homology
of various complexes, including weighted cubical complexes.
We computed persistent homology of two 3D greyscale images, 
Bonasai and Head Aneurism\footnote{available at \url{https://klacansky.com/open-scivis-datasets/}}, 
and the greyscale Lena 2D image
 (see Table \ref{tab:data}). 
We converted the data type of the datasets to double precision (64bit floating-point numbers).
To see the scalability, halved and quartered images for Bonsai and Head Aneurism,
and doubled and quadrupled images for Lena were also used.
The computation time and memory usage are shown in Tables \ref{tab:timing} and \ref{tab:memory}.
The timing results include the loading time of the program itself and file IO, which may be dominant for small images.

Note that due to insufficient memory, DIPHA could not finish the computation for Head Aneurism $512^3$
on our machine (see Table \ref{tab:machine}).
On the other hand, with sufficient memory and many cores, DIPHA would outperform Cubical Ripser.

\begin{table}[ht]\scriptsize
\begin{tabular}{|c|c|}
\hline
CPU & AMD Ryzen Threadripper 2990WX 32-Core Processor \\
\hline
Memory & 64GB \\
\hline
OS & Ubuntu 18.04.4 LTS \\
\hline
Compiler & GCC 7.5.0 \\
\hline
MPI (for DIPHA) & Intel MPI 3.3.2 \\
\hline
\end{tabular}\caption{Machine configuration used for experiments}\label{tab:machine}
\end{table}

\begin{table}[ht]\scriptsize
\begin{tabular}{|c|c|c|c|c|c|c|c|c|c|}
\hline
& Lena512 & Lena1024 & Lena2048 & Bonsai64 & Bonsai128 & Bonsai256 & Head128 & Head256 & Head512 \\
\hline
Size & $512^2$ & $1024^2$ & $2048^2$ & $64^3$ & $128^3$ & $256^3$ & $128^3$ & $256^3$ & $512^3$ \\
\hline
0-cycles & 17789 & 29382 & 35559 & 2015 & 8207 & 13751 & 168857 & 550014 & 703506 \\
1-cycles & 10253 & 21079 & 28509 & 5871 & 25117 & 46777 & 231250 & 858765 & 1497812 \\
2-cycles & 0 & 0 & 0 & 3026 & 14898 & 32300 & 48537 & 213409 & 625466 \\
\hline
\end{tabular}\caption{Dataset specification: 
image size and the number of cycles.
}\label{tab:data}
\end{table}


\begin{table}[ht]\scriptsize
\begin{tabular}{|c|c|c|c|c|c|c|c|c|c|c|}
\hline
& Lena512 & Lena1024 & Lena2048 & Bonsai64 & Bonsai128 & Bonsai256 & Head128 & Head256 & Head512 \\
\hline
Cubical Ripser (single process)&  0.18 & 0.81 & 3.3 & 0.6 & 4.5 & 38 & 6.6 & 50 & 355 \\
DIPHA (single process)& 0.71 & 2.8 & 11 & 2 & 13 & 106 & 15 & 123 & NA \\
DIPHA-d (single process) & 0.74 & 3.0 & 12 & 2.3 & 16 & 128 & 16.7 & 143 & NA \\
DIPHA (MPI 2 process) & 0.45 & 1.7 & 7.4 & 1.7 & 9.8 & 79 & 10.5 & 86 & NA \\
DIPHA (MPI 4 process) & 0.36 & 0.98 & 4.4 & 1.4 & 6.8 & 59 & 6.5 & 61 & NA \\
DIPHA (MPI 8 process) & 0.40 & 0.88 & 3.0 & 1.1 & 5.3 & 44 & 4.1 & 38 & NA \\
\hline
\end{tabular}\caption{Timing measured in seconds (5-fold average). DIPHA-d uses the dualization (cohomology) algorithm.
NA means the computation did not finish due to insufficient memory}\label{tab:timing}

\begin{tabular}{|c|c|c|c|c|c|c|c|c|c|c|}
\hline
& Lena512 & Lena1024 & Lena2048 & Bonsai64 & Bonsai128 & Bonsai256 & Head128 & Head256 & Head512 \\
\hline
Cubical Ripser & 35MB & 127MB & 494MB & 55MB & 414MB & 3.1GB & 443MB & 3.1GB & 22GB \\
DIPHA & 112MB & 380MB & 1.4GB & 208MB & 1.4GB & 11GB & 1.4GB&  11GB&  NA \\
DIPHA-d & 113MB	&380MB&	1.4GB&	206MB	&1.4GB	&11GB&	1.4GB&	11GB	&NA \\
\hline
\end{tabular}\caption{Memory usage}\label{tab:memory}
\end{table}

\begin{rem}
Cubical Ripser has a command-line option (\verb+--min_cache_size+) to control the trade-off between speed and memory usage
by suppressing caching.
For example, for the Bonsai256 dataset \\
\verb+--min_cache_size 5000+ takes 60s and 2.5GB memory
against the default (\verb+--min_cache_size 0+) taking 38s and 3.1GB memory.
However, the memory gain does not usually pay for the much longer computation time.
\end{rem}

\begin{rem}
After the first preprint of this note was made public,
we were urged by one of the developers of the Topology ToolKit (TTK)\footnote{\url{https://github.com/topology-tool-kit/ttk}, we tested with the version ``commit 11fd73d''.}
that we should mention and perform a comparison with TTK,
which was claimed to offer a faster computation of persistent homology of images/volumes.
However, when we carefully examined the program, we found a serious flaw of TTK.
For example, TTK reports that the following $3\times 3$-image
\[
\begin{pmatrix}
0 & 0 & 0 \\
0 & 9 & 0 \\
0 & 0 & 0
\end{pmatrix}
\]
has no $1$-cycle.
We reported the above example to the developer, pointing out a possible cause why TTK failed to find the 1-cycle\footnote{Alexander duality is possibly used in a wrong manner.}.
The developer simply rejected our argument, insisting that TTK's computation was correct according
 to his definition of persistent homology.
 There is no account in the literature of the definition nor the algorithm for persistent homology used in the TTK.
What we have understood is that TTK first triangulates the image and computes with a certain weighted simplicial complex.
Thus, the results may naturally be different from those of Cubical Ripser, which computes persistent homology of the cubical complex associated with the image.
However, we believe that the above image should have a 1-cycle with a significant (nine, in most cases) lifetime
no matter what complex is associated with it.
For this reason, we did not choose TTK as our comparison target, but we chose the well-established program DIPHA.
We note that the outputs of Cubical Ripser and DIPHA coincide.
We would like to emphasise that overall TTK is a wonderful piece of software with various functionality.
We hope that the definition and the algorithm for persistent homology used in TTK will be clarified in the near future.
\end{rem}

\section{Example}\label{sec:example}
In this section, we give a quick instruction on how to use our software
for image analysis.

The python module of Cubical Ripser is easily installed with
\begin{itembox}\small
\begin{verbatim}
> pip install git+https://github.com/shizuo-kaji/CubicalRipser_3dim 
\end{verbatim}
\end{itembox}
Also one can easily build the program from the source available at \cite{github} on any machine with C++11 compilers such as G++, Clang, or Microsoft C++.

\begin{ex}[JPEG image]
Given an JPEG image \verb+input.jpg+,
we first convert it into a 2D Numpy array \verb+input.npy+:
\begin{itembox}\small
\begin{verbatim}
> python demo/img2npy.py input.jpg input.npy 
\end{verbatim}
\end{itembox}

Code \ref{ph} computes the persistent homology of the image with Python.
\begin{lstlisting}[caption={Python codes for computing persistent homology},label=ph]
# import the Numpy module
import numpy as np
# import the Cubical Ripser python module
import cripser
# load the image in the numpy array format
arr = np.load("input.npy").astype(np.float64)
# compute the persistent homology up to degree 1
result = cripser.computePH(arr,maxdim=1,location="birth")
\end{lstlisting}
Here, \verb+result+ is another 2D Numpy array of shape $(M,6)$,
where $M$ is the number of cycles.
The six numbers of each row indicate
the dimension of the cycle, birth-time, death-time, and location ($x,y,z$) of the cell giving birth to the cycle (see Example \ref{ex:location}).

Alternatively,
one can use the command-line executable
to compute the persistent homology
of the 1D/2D/3D Numpy array \verb+input.npy+
and obtain results in \verb+result.csv+.
\begin{itembox}\small
\begin{verbatim}
> ./cubicalripser --location birth --output result.csv input.npy
\end{verbatim}
\end{itembox}
Each line in the output \verb+result.csv+ consists of six numbers indicating
the dimension of the cycle, birth-time, death-time, and location ($x,y,z$). 
\end{ex}

\begin{ex}[3D volume image in DICOM]
Given a series of DICOM files named \verb+input00.dcm+, 
\verb+input01.dcm+, \verb+input02.dcm+...,
one can convert them to a single 3D Numpy array
\verb+volume.npy+ 
that is compatible with Cubical Ripser by
\begin{itembox}\small
\begin{verbatim}
> python demo/img2npy.py input*.dcm volume.npy 
\end{verbatim}
\end{itembox}
A series of image files such as JPEG and PNG 
(as long as the Pillow library can handle them)
can also be made into a volume in a similar way.

Note that here we rely on the shell's path expansion.
If your shell does not support it,
you can manually specify file names as in the following:
\begin{itembox}\small
\begin{verbatim}
> python demo/img2npy.py input00.dcm input01.dcm input02.dcm volume.npy 
\end{verbatim}
\end{itembox}
\end{ex}

\begin{ex}[Preprocessing]
There are typically two ways to obtain an image from the observed data.
Suppose we have a greyscale picture. 
If the features we are interested in is related to its pixel intensity values, 
we would use the greyscale picture as it is so that $\phi(x,y,z)$ is the intensity of the pixel at $(x,y,z)$.
On the other hand, if we are more interested in its spatial pattern, 
we may want to encode spatial information into the function value of $\phi$.
One popular way to do so is to use the distance transform.
First, we make the greyscale picture binary (black and white) by \emph{thresholding}.
A drawback is that we have to choose how to threshold (see \cite{threshold}).
Then, apply the \emph{distance transform}, 
which assigns each pixel with the signed distance to the nearest boundary between black and white (see Figure \ref{fig:DT}).
This process is easily done with the SciPy \cite{scipy} and the Scikit-Image Python package \cite{Scikit} as described in Code \ref{dt}.

\begin{figure}[ht]
\center
\includegraphics[width=8cm]{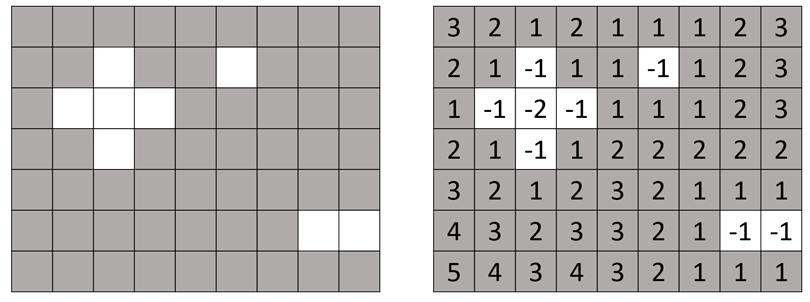}
\caption{A binary image (left) and its distance transform with respect to the $L^1$-norm (right) }\label{fig:DT}
\end{figure}

\begin{lstlisting}[caption={Python codes for thresholding and distance transform},label=dt]
# img is a 2D or 3D Numpy array 
from scipy.ndimage.morphology import distance_transform_edt
from skimage.filters import threshold_otsu
bw_img = (img >= threshold_otsu(img))
dt_img = distance_transform_edt(bw_img)-distance_transform_edt(~bw_img)
\end{lstlisting}
\end{ex}

\begin{ex}[Scalar time-series]
A scalar time-series can be considered as a 1D image,
so Cubical Ripser can compute its persistent homology.
Note that other software would be more efficient for this purpose.
We just give a toy example here\footnote{
A little more involved example of regressing the frequency of noisy sine curves
is demonstrated at \url{https://github.com/shizuo-kaji/TutorialTopologicalDataAnalysis}}.
.
\begin{itembox}\small
\begin{verbatim}
# Python script for a 1D example
import numpy, cripser
x = numpy.linspace(0,4*numpy.pi,100)
img = numpy.sin(x)+0.1*x
pd = cripser.computePH(img)
\end{verbatim}
\end{itembox}

\end{ex}

\begin{ex}[Data exchange with DIPHA]
We included a little script, \verb+dipha2npy.py+, to convert input and output files
 between Cubical Ripser and DIPHA.
\begin{itembox}\small
\begin{verbatim}
# convert an Numpy array "img.npy" into DIPHA's format "img.complex"
> python dipha2npy.py img.npy img.complex 
# the other way around
> python dipha2npy.py img.complex img.npy
# convert DIPHA's output "result.output" into an Numpy array "result.npy"
> python dipha2npy.py result.output result.npy 
\end{verbatim}
\end{itembox}
\end{ex}

\begin{ex}[Localised features]\label{ex:location}
Homology encodes global information, but it contains certain local information as well.
Cubical Ripser outputs for each homological feature (cycle) the location of the cell where
 it is born (or optionally, killed)\footnote{A small remark is that the location is not uniquely determined, and the program chooses a representative.}.
These coordinates can be used to localise roughly where the homological features exist.
We can think of this information as a kind of annotation to the input image which can be utilised in conjunction with
other image analysis techniques 
(Example \ref{ex:NN}).

For an input image,  
we make another image for each $d=0,1,2$ whose pixel value is the maximum lifetime (death-time minus birth-time)
of the cycle at the location. 
This image is usually very sparse with a lot of zeros.
By stacking the original image and these images for $d=0,1,2$, we obtain a 4-channel image, which we call the \emph{lifetime enhanced image}.
The lifetime enhanced image can be obtained  as follows:
\begin{itembox}\small
\begin{verbatim}
# Compute persistent homology of the image input.npy
> ./cubicalripser --location birth --output result.npy input.npy
# From the output result.npy, create the lifetime enhanced image 
> python demo/stackPH.py result.npy -o lifetime_image.npy -i input.npy
\end{verbatim}
\end{itembox}
See Figure \ref{fig:stacked} for a 2D image example.

Another way to incorporate persistent homology information to the original image is to keep 
the histogram of the homological feature for each pixel;
for each pixel, make the 2D histogram of (birth, lifetime) of the cycles located at the pixel.
Then, flatten the histogram to obtain a vector for each pixel.
The resulting image has the same number of channels as the dimension of the flattened vector of the histogram.
We call the image the \emph{persistent histogram image}.
The script \verb+stackPH.py+ produces the persistent histogram image
if executed with the command-line option ``\verb+-t hist+''.
In this way, we can encode more information than lifetime.
The drawback is that the resulting stacked image has many channels and requires
more computational resource.
Persistent histogram image can be considered as a local variant of the persistence image \cite{persistence_images}
which provide a vector representation of persistent homology\footnote{We do not apply a kernel density estimator as persistence image does because if we feed the persistent histogram image into a convolutional neural network, the first layer is expected to learn the optimal kernel from data.}.
\end{ex}

\begin{ex}[Combination with Neural Networks]\label{ex:NN}
A neural network with many layers trained with a lot of data would successfully learn global features of images
by piling up local information.
However, when data is scarce (e.g., in medical and civil engineering applications), 
it is beneficial to help neural networks by feeding them with mathematically defined global features.
For this purpose, we can use the lifetime enhanced images
considered in Example \ref{ex:location} as input for convolutional neural networks to achieve specific tasks.
One of the advantages of this method is that we do not have to modify the architecture of the existing neural networks
except for the number of input channels. This means, if we have a neural network to do a specific task,
we can try this method very easily with a minimal change.

To demonstrate this idea, we conducted an experiment of the classification task of the Reduced MNIST dataset \cite{rmnist}.
The dataset consists of the MNIST handwritten digits $0$ through $9$, but the training dataset contains only 10 images,
one sample from each class. The test dataset contains 10000 images.
Our task is to classify the images into ten categories.
The original images are $28\times 28$ greyscale.
We scaled them up to $56\times 56$ so that it gets more difficult for a convolutional neural network to grasp
global information.

We used a simple convolutional neural network
 with three Conv-BatchNorm-ReLU-MaxPool-Dropout layers followed by a fully-connected layer.
 We fed them with two types of inputs: just the images of digits (denoted by \emph{without PH})
and the lifetime enhanced images
of $d=0,1$ computed from the distance transform (denoted by \emph{with PH}).
The lifetime, in this case, corresponds roughly to the size of the feature such as loops in the images.
An example of the lifetime enhanced images is shown in Figure \ref{fig:stacked}.

\begin{figure}[ht]
\center
\includegraphics[width=14cm]{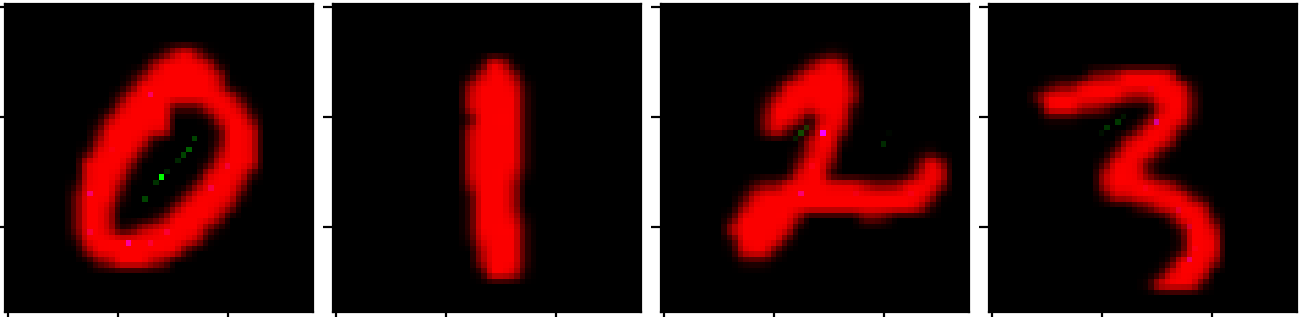}
\caption{Lifetime enhanced image: the original greyscale images is displayed as the red channel,
the $d=0$ lifetime image is displayed as the green channel,
and the $d=1$ as the blue channel.}\label{fig:stacked}
\end{figure}

We measured the accuracy and the top-2 accuracy (the network can make two predictions for each image).
The results are shown in Figure \ref{fig:mnist}.

\begin{figure}[ht]
\center
\includegraphics[width=7cm]{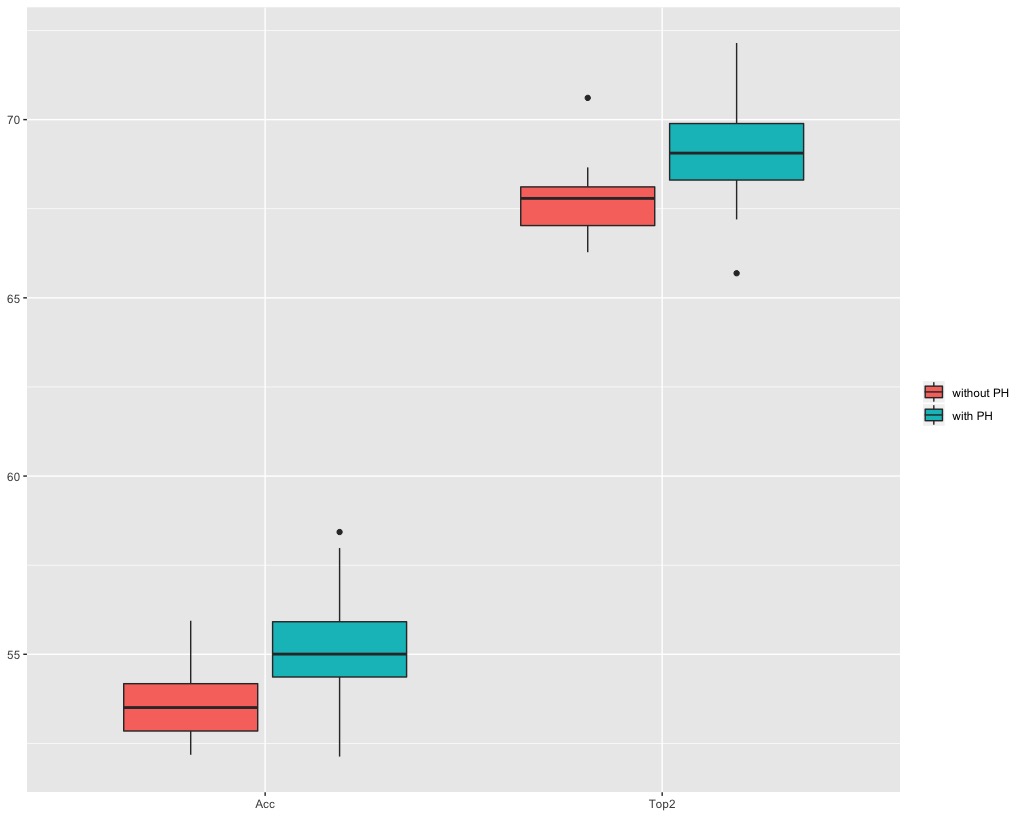}
\includegraphics[width=9cm]{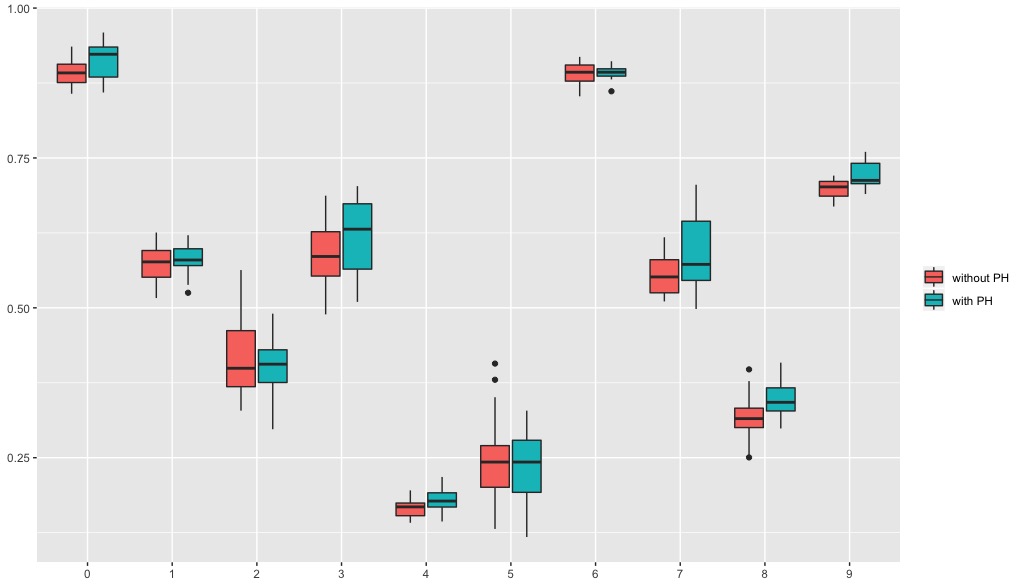}
\caption{Reduced MNIST classification results: (Left) accuracy and Top-2 accuracy measured twenty times, (Right) breakdown for each class}\label{fig:mnist}
\end{figure}

From the boxplot of twenty measurements for each configuration,
we observe that the lifetime information generally increases the performance 
although the performance fluctuates due to minimal training data.
We note that the performance gain for those ``topologically conspicuous'' figures such as $0,8$, and $9$ are relatively large.

We conducted similar experiments with the 3D MNIST dataset\footnote{\url{https://www.kaggle.com/daavoo/3d-mnist}},
which consists of volumetric images of digits converted from MNIST.
With the 3D MNIST, we did not see any significant increase in performance. This may be attributed to the fact that 
topological features such as holes are wiped out in the images
due to the low resolution ($16 \times 16\times 16)$.

The results and the codes for both experiments are made available\footnote{\url{https://github.com/shizuo-kaji/HomologyCNN}}
in the hope that they provide a hands-on example for applying our method to other image analysis tasks.
\end{ex}


\section{Acknowledgements}
We thank Ippei Obayashi, who is the author of the program ``\emph{homccube}'' for fast computation of persistent homology of images, 
for his useful comments.
We also thank Julien Tierny for providing us with the information of the Topology ToolKit.


\end{document}